\documentclass[conference]{IEEEtran}
\IEEEoverridecommandlockouts
\usepackage{color,soul}
\usepackage{hyperref}
\usepackage{url}
\usepackage{algorithm}
\usepackage{algorithmic}
\usepackage{subcaption}
\usepackage{booktabs}
\usepackage{graphicx}
\usepackage{subcaption}

\usepackage{float}
\usepackage{cite}
\usepackage{amsmath,amssymb,amsfonts}
\usepackage{algorithmic}
\usepackage{graphicx}
\usepackage{textcomp}
\usepackage{xcolor}
\def\BibTeX{{\rm B\kern-.05em{\sc i\kern-.025em b}\kern-.08em
    T\kern-.1667em\lower.7ex\hbox{E}\kern-.125emX}}
\begin{document}

\title{Leveraging Relational Information for Learning Weakly Disentangled Representations
}

\author{\IEEEauthorblockN{Andrea Valenti}
\IEEEauthorblockA{\textit{Department of Computer Science} \\
\textit{University of Pisa}\\
Pisa, Italy \\
andrea.valenti@phd.unipi.it}
\and
\IEEEauthorblockN{Davide Bacciu}
\IEEEauthorblockA{\textit{Department of Computer Science} \\
\textit{University of Pisa}\\
Pisa, Italy \\
davide.bacciu@unipi.it}
}

%\author{(AUTHORS' INFORMATION REDACTED FOR DOUBLE BLIND REVIEW)}

\maketitle

\begin{abstract}
Disentanglement is a difficult property to enforce in neural representations. This might be due, in part, to a formalization of the disentanglement problem that focuses too heavily on separating relevant factors of variation of the data in single isolated dimensions of the neural representation. We argue that such a definition might be too restrictive and not necessarily beneficial in terms of downstream tasks. In this work, we present an alternative view over learning (weakly) disentangled representations, which leverages concepts from relational learning. We identify the regions of the latent space that correspond to specific instances of generative factors, and we learn the relationships among these regions in order to perform controlled changes to the latent codes. We also introduce a compound generative model that implements such a weak disentanglement approach. Our experiments shows that the learned representations can separate the relevant factors of variation in the data, while preserving the information needed for effectively generating high quality data samples.
\end{abstract}


\section{Introduction}
While trying to find a way to reproduce aspects of natural intelligence into artificial systems, researchers proposed the notion of \emph{meta-priors}, first introduced by \cite{2013_bengio_representation} and then further refined and expanded in \cite{2018_tschannen_recent_advances}. A meta-prior is a generic assumption about the world that is expected to hold true for all possible tasks that an artificial agent might encounter in the future, thus providing a way to structure the learned representations in a useful way for possible downstream tasks. In the latest years, meta-priors have helped the representations learned by neural networks to reach levels of expressivity that were unthinkable just a few decades ago. Modern distributed representations can, for instance, disentangle factors of variations of the data, encode hierarchical features at different levels of abstractions, express the natural clustering organization of the data, and incorporate various types of supervised information \cite{2018_tschannen_recent_advances}.

Despite these achievements, finding a way for reliably enforcing different kinds of meta-prior is still an open research question. In particular, one of the most difficult meta-prior to impose on the learned representations is \emph{disentanglement}. One of the challenges that arises when dealing with the disentanglement problem, is that a formal definition of what constitute a disentangled representation is still a matter of debate \cite{do2019theory}. Many works just assume that a disentangled representation is a representation in which each latent dimension is responsible for encoding a single generative factor of the data. We argue that this intuitive definition can be too strict in general, as it is possible for distinct factors of variations to manifest themselves in the data only in an entangled way. Sometimes, only a subset of all the possible factors of variations is worth disentangling, while the others can be left entangled. Moreover, it has been shown that the imposition of this form of disentanglement on the learned representations can actually damage the overall performance on downstream tasks, instead of providing a clear benefit \cite{2019_locatello_challenging}.

For these reasons, in this work we wish to introduce a different approach on disentanglement and disentangled representations, which we call \emph{weak disentanglement}. A weakly disentangled representation is a representation where the generative factors are not encoded into specific separate dimensions. The information about the original values of generative factors is instead encoded into different regions of the latent space, with each region identifying a specific combination of factors. Given a weakly disentangled representation, it is therefore possible to recover the original generative factors by checking in which region of the latent space that representation ends up. %getting encoded.

In particular, we propose a new generative neural model for the learning of weakly disentangled representations. The main components of this model are the Abstraction Autoencoder (AbsAE) and the Relational Learner (ReL). The AbsAE is an adversarial autoencoder \cite{2015_makhzani_adversarial} augmented with an adaptive prior distribution that is able to identify the regions of the latent space containing the relevant instances of generative factors, using only a minimal amount of supervised information. During training, the ReL learns how to navigate such structured latent space, moving the input representations into new regions of the same latent space according to a set of predefined relations. These modules, together, are able to learn representations that, while still being entangled from a ``classic" point of view, allow for being easily manipulated in order to induce controlled changes on the chosen factors of variations. In the rest of this paper, we will show that this form of weak disentanglement can obtain representations that preserves all the relevant information for reconstructing the original data, while at the same time allowing for the manipulation of one or more factors of variations in a compositional way. The main contribution of this paper are the following:
\begin{enumerate}
    \item We introduce the new practical notion of \emph{weak disentanglement}.
    \item We introduce a new generative model and training procedure for effectively learning weakly disentangled representations.
    \item We show how relational information can be used to induce a prior distribution over the latent space of the model, useful for the weak disentanglement task.
\end{enumerate}

\begin{figure*}
     \centering
     \begin{subfigure}{.495\textwidth}
         \centering
         \includegraphics[width=\textwidth]{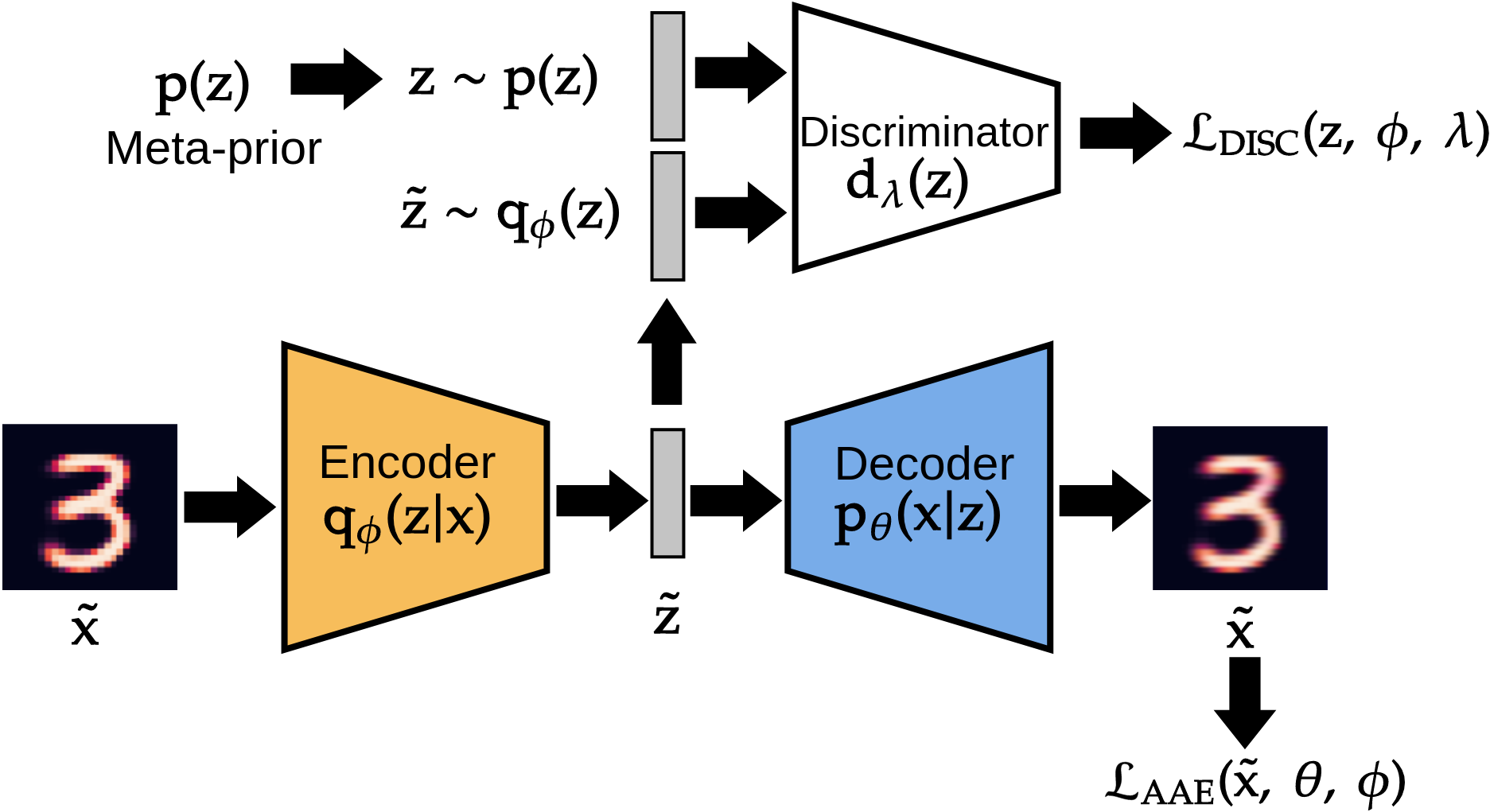}
         \caption{(AbsAE): the encoder and the decoder are trained to learn the mapping between the data space and the latent space. The latent codes are forced to follow the prior distribution $p(z)$.}
         \label{fig_arch_absae}
     \end{subfigure}
     %%\fill
     \begin{subfigure}{.495\textwidth}
         \centering
         \includegraphics[width=\textwidth]{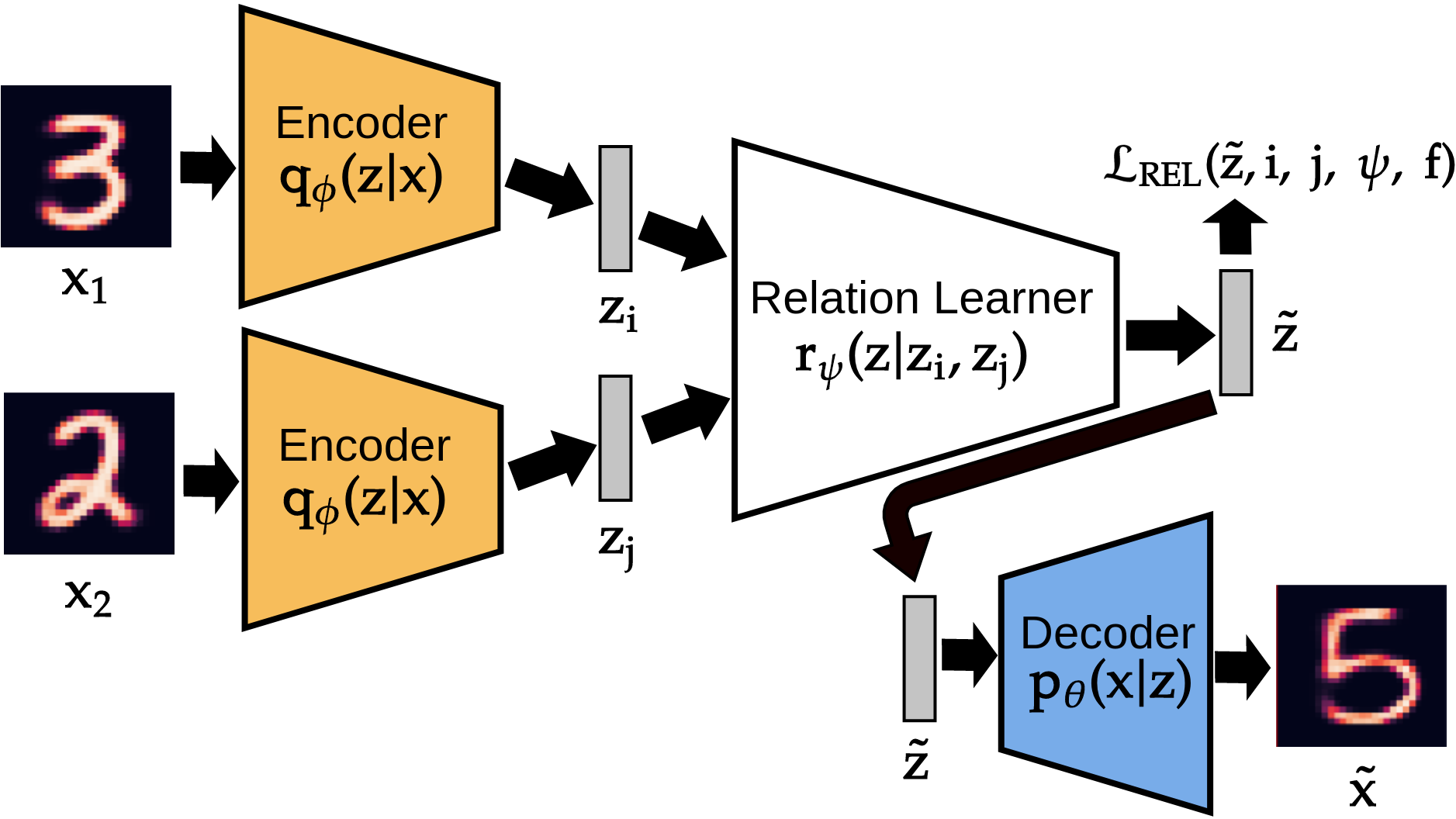}
         \caption{(ReL): The relation learner is trained in the latent space. The encoder and the decoder are the same as AbsAE.}
         \label{fig_arch_rel}
     \end{subfigure}
        \caption{Overall proposed architecture, composed of the Abstraction Autoencoder (AbsAE) and the Relational Learner (ReL).}
        \label{fig_architecture}
\end{figure*}

\section{Related Works}
\label{sec_related_works}

After the inital introduction of the concept of meta-priors \cite{2013_bengio_representation, 2018_tschannen_recent_advances}, many works explored different ways to enforce meta-priors on the representations learned by autoencoders.
It is possible to identify approximately three main approaches: i) Using regularization constraints on the encoder's posterior distribution $q_\phi(z|x)$, ii) Using architectural constraints on either the encoder $q_\phi(z|x)$ or the decoder $p_\theta(x|z)$ (or both) and iii) Choosing flexible prior distributions $p(z)$. Many works combine these approaches in order to force specific properties on the learned representations.

Focusing on disentanglement, the early methods are mainly concerned on re-weighting the second term of the ELBO, such as the $\beta$-VAE \cite{2017_higgins_beta-vae}, and the $\beta$-VAE-2 \cite{2018_burgess_understanding}. The main shortcoming of this approach is that disentanglement is achieved at the expense of reconstruction accuracy, hampering the performance on subsequent downstream tasks.
Another line of works builds upon the ELBO decomposition provided by \cite{2016_hoffman_elbo} to separately penalize different terms, such as FactorVAE \cite{2018_kim_factor_vae}, $\beta$-TCVAE \cite{2018_chen_beta-TCVAE}, InfoVAE \cite{2019_zhao_infovae} and DIP-VAE I \& II \cite{2017_kumar_variational}. They all apply several weighting factors on different parts of the ELBO in order to emphasize specific properties on the latent representations. For example, \cite{2018_kim_factor_vae} and \cite{2018_chen_beta-TCVAE} try to achieve disentanglement by encouraging the total correlation between the latent dimensions to be as low as possible. 
Other works such as HSIC-VAE \cite{2018_lopez_hsic-vae} and HFVAE \cite{2019_esmaeili_structured} try to enforce independence between groups of latent variables. While being able to isolate simple generative factor to some degree, in general such models struggles to achieve a reasonable disentanglement when the factor of variation cannot be identified by straightforward mathematical notions such as statistical independence (which is often the case with real-world data). 

%PixelGAN-AE \cite{2017_makhzani_pixelgan} and the Variational Information Bottleneck method \cite{2016_alemi_deep,2018_achille_information} try to prevent the tendency of the latent code to be ignored when the decoder is too powerful. 

An interesting line of works tries to leverage different degrees of supervised information in order to achieve disentanglement \cite{2015_louizos_variational, 2015_kulkarni_dc-ign, 2017_lample_fader, 2019_locatello_few-labels, gabbay2021image}. For example, \cite{2019_locatello_few-labels} relies on complete supervision of a small subset of training data. \cite{gabbay2021image} further relaxs these constraints by requiring only partial annotations on a subset of generative factors. However, the lack of a rich-structure in the latent space makes it impossible to associate a confidence level to the models' predictions. Some works in the field of concept learning also try to build a structured latent space distribution in order to isolate relevant high-level concepts associated with the data samples \cite{rostami2020concept} \cite{koh2020concept}. In particular, \cite{rostami2020concept} use an additional classification network on the latent space in order to cluster together representations associated to the same concept in a continual learning setting. In \cite{hosoya2018group}, the authors strongly disentangle the latent space into group-common ``content" variables and instance-specific ``transformation" variables. These approaches are suitable for identifying the different values of a single relevant generative factor (i.e. the concept, or the data group), but cannot be applied when more than one factor needs to be disentangled from the data.

Finally, there exists a few works that focus on leveraging relational information among data samples. \cite{locatello2020without_compromises} uses pairs of images where the value of a random subset of generative factors is different. On the other hand, \cite{chen2020pairwise_similarity} uses weak supervision between a pair of images consisting in a similarity score about a factor to be disentangled.
 \cite{bai2021contrastively} strongly disentangle representation of sequential data into ``static" factors, that are constant along all the duration of the sequence, and ``dynamic" factors, that vary across the timesteps.
These approaches, while powerful in principle, typically require the training data to be structured in very specific ways in order to be able to make use of the supervised information available. This can limit their application to a wide variety of tasks.

\section{Weak Disentanglement of Latent Representations}
\label{sec_methods}
The section introduces our approach  implementing the \emph{weak disentanglement} meta-prior. A schematic view of the model's architecture is shown in Figure \ref{fig_architecture}. The model loss is trained using the following objective:
\begin{equation}
\label{eq_overall_loss}
    \mathcal{L} = \mathcal{L}_{AE}(x, \theta, \phi) + \beta \mathcal{L}_{DISC}(z, \phi, \lambda) +  \gamma \mathcal{L}_{REL}(z, \psi, f),
\end{equation}
where $\theta$, $\phi$, $\lambda$ and $\psi$ are the parameters of the encoder, decoder, discriminator and relational learner respectively, $x$ is a data sample and $z$ is its corresponding latent representation, $f$ is the particular relation that we would like to learn.
The hyperparameters $\beta$ and $\gamma$ are used to adjust the importance of the different terms.
The first two terms of Eq.\ref{eq_overall_loss} correspond to the auto-encoding part of the model, while the last term is the relational part.
The AbsAE's task is to learn a mapping from the data space to an abstract latent space, and vice-versa. The latent space is encouraged to follow a specific \emph{meta-prior distribution $p(z)$}, an adaptive gaussian mixture (GM) distribution built ad-hoc for the task. This distribution is able to clusters different instances of the same generative factors into a similar region of the latent space. At the same time, the ReL is trained to learn relations between such regions of the latent space, exploiting the prior distribution $p(z)$ learned by the AbsAE. We alternate a training iteration of the AbsAE with a training iteration of the ReL in order to gradually learn both the relevant combinations of generative factors and the relations between them. In the rest of this section, we provide a detailed description of each of the modules.

\subsection{Abstraction Autoencoder}
\label{sec_absae}
The AbsAE, depicted in Figure \ref{fig_arch_absae}, is composed of two sub-networks: the \emph{encoder} $q_{\phi}(z|x)$ and \emph{decoder} $p_{\theta}(x|z)$, parameterized by  $\phi$ and $\theta$, respectively. The mapping between data space and latent space is learned by optimizing the following maximum likelihood objective:
\begin{align}
\label{eq_aae_obj}
\max_{\theta, \phi}& \quad \mathcal{L}_{AE}(x, \theta, \phi) + \beta \mathcal{L}_{DISC}(z, \phi) = \\
\label{eq_aae_obj_2}
&= \mathbb{E}_{q_\phi(z|x)} \left[ \log p_\theta(x|z) \right] - \beta D\left( q_\phi(z) || p(z) \right),
\end{align}
where $D$ is an arbitrary divergence (such as the Kullback-Leibler divergence) and $\beta$ is an hyperparameter controlling the amount of desired regularization. The first term of Eq.\ref{eq_aae_obj_2} encourages the latent codes $z$ to be an informative representation of the corresponding original input $x$, while the second term encourages $z$ to follow a desired prior distribution $p(z)$. Since this second term is  generally not computable for an arbitrary choice of $q_\phi(z)$ and $p(z)$\cite{2017_mescheder_adversarial}, we estimate it using an additional \emph{discriminator} network $d_\psi(z)$, parameterized by $\lambda$. Thus, the second term of Eq. \ref{eq_aae_obj_2} is optimized in an adversarial way via the following objective
\begin{align}
\min_\phi \max_\lambda & \quad \mathcal{L}_{DISC}(z, \phi, \lambda) = \\
    &= \mathbb{E}_{q_\phi(z)}[ \log d_\lambda(z)] + \mathbb{E}_{p(z)}[\log(1-d_\lambda(z))].
\end{align}
The flexibility introduced by the adversarial estimation of the divergence $D$ allows us to chose any distribution $p(z)$ that best fits our needs. Since our goal is to identify and disentangle the different combinations of generative factors that appear in the data, we choose the prior distribution $p(z)$ to be a a gaussian mixture (GM) 
\begin{equation}
    \label{eq_prior}
    p(z) = \frac{1}{N}\sum_{i=1}^{N} p_i(z) = \frac{1}{N}\sum_{i=1}^{N} \mathcal{N}(\mu_{i}, \Sigma^2_i),
\end{equation}
where $N$ is the number of the distinct factor combinations in the data. The mean $\mu_i$ and the covariance $\Sigma^2_i$ of each prior component are estimated empirically from a small subset of supervised samples, containing ancillary information that describes properties of the data that are relevant for the relations we wish to learn
(for practical examples of generative factor values see Section \ref{sec_datasets}):
\begin{equation}
    y = \left\{ (y_{g_1}, y_{g_2}, ..., y_{g_K}) \right\}_i^{N_y},
\end{equation}
where $g_i$ is the $i$-th generative factor, $y_{g_i}$ is the label associated to $g_i$, $K$ is the total number of factors and $N_y$ is the number of labelled data.
We leverage the fact that the auto-encoding training procedure of AbsAE tends to naturally organize the latent space in an efficient way, with similar samples (i.e. samples for which the generative factors have the same value) that ends up to be encoded in the same region of the latent space. The labelled samples are a way to identify the relevant regions of interest, and the prior distribution $p(z)$ helps into shaping those regions into a GM distribution that is easy to model and manipulate in a meaningful way.
This setting grants a high flexibility on what the relevant factors can be. For example, it is possible to specify only a subset of such factors of variations, so that the remaining factors will be treated as nuisances.

\begin{table*}
   \centering
   \caption{Latent space classification accuracy of the AbsAE on the HWF, dSprites and Shapes3D datasets. ACC denotes the accuracy, AR the accepted ratio. $\tau$ is the supervision amount, measured in number of samples.}
   \label{tab_clustering}
    \begin{tabular}{lccccccc}
        \toprule
         \multicolumn{2}{c}{} &
         \multicolumn{2}{c}{\bf HWF} & \multicolumn{2}{c}{\bf dSprites} & \multicolumn{2}{c}{\bf Shapes3D} \\
         $\alpha$ & $\tau$ & ACC & AR & ACC & AR & ACC & AR \\
         \cmidrule(rl){3-4}\cmidrule(rl){5-6}\cmidrule(rl){7-8}
        
        Previous work \cite{li2020closed} & -- & 0.997 & 1.0 & -- & -- & -- & -- \\
        \midrule
        AbsAE, $\alpha$=0.0 & 10  & 0.917  & 1.0  & 0.548  & 1.0  & 0.115  & 1.0 \\
         & 20  & 0.977 & 1.0 & 0.554 & 1.0 & 0.272 & 1.0 \\
         & 30  & 0.982 & 1.0 & 0.590 & 1.0  & 0.370  & 1.0 \\
        \midrule
        AbsAE, $\alpha$=0.1 & 10 & 0.954  & 0.995  & 0.592  & 1.0  & 0.220 & 0.749  \\
         & 20 & 0.992 & 0.918 & 0.631 & 0.999 & 0.478  & 0.811 \\
         & 30 & 0.987 & 0.995 & 0.642 & 1.0  & 0.553 & 0.873 \\
        \midrule
        AbsAE, $\alpha$=0.3 & 10 & 0.978 & 0.988  & 0.676  & 0.991  & 0.617 & 0.752\\
         & 20 & 0.990 & 0.982 & 0.700 & 0.993 & 0.713 & 0.780\\
         & 30 &  1.0  & 0.985 & 0.728 & 0.993 & 0.779 & 0.802 \\
        \midrule
        AbsAE, $\alpha$=0.5 & 10 & 0.983  & 0.973  & 0.865 & 0.988  & 0.699  & 0.753 \\
         & 20 & 0.993 & 0.974 & 0.882 & 0.990 & 0.853 & 0.714\\
         & 30 & 0.998 & 0.982 & 0.891 & 0.989 & 0.890  & 0.781 \\
        \midrule        
        AbsAE, $\alpha$=0.7 & 10 & 0.982  & 0.970  & 0.948  & 0.994  & 0.821  & 0.673 \\
         & 20 & 0.995 & 0.971 & 0.937 & 0.989 & 0.881 & 0.738\\
         & 30 & 0.994 & 0.986 & 0.966 & 0.993 & 0.902 & 0.759 \\
        \midrule
        AbsAE, $\alpha$=0.9 & 10 & 0.993  & 0.957  & 0.921  & 0.981  & 0.854 & 0.646 \\
         & 20 & 0.999 & 0.953 & 0.956 & 0.987 & 0.893 & 0.758\\
         & 30 & 0.999 & 0.951 & 0.976 & 0.985 & 0.910 & 0.711 \\
        \bottomrule
    \end{tabular}
\end{table*}

\begin{table*}
    \centering
    \caption{Relational accuracy of the ReL on the HWF, dSprites and Shapes3D datasets. ACC denotes the accuracy, AR the accepted ratio.}
    \label{tab_relational}    
    \begin{tabular}{lccccccc}
        \toprule
        \multicolumn{2}{c}{} & \multicolumn{2}{c}{\bf HWF} & \multicolumn{2}{c}{\bf dSprites} & \multicolumn{2}{c}{\bf Shapes3D} \\
        $\alpha$ & Depth & ACC & AR & ACC & AR & ACC & AR \\
                 \cmidrule(rl){3-4}\cmidrule(rl){5-6}\cmidrule(rl){7-8}
        Previous work \cite{li2020closed} & 1 & 0.985 & 1.0 & -- & -- & -- & -- \\
        \midrule
        ReL, $\alpha$=0.0 & 1 & 0.9966 & 1.0 & 0.9896 & 1.0 & 0.7521 & 1.0\\
                     & 5 & 0.9939 & 1.0 & 0.9898 & 1.0 & 0.7378 & 1.0\\
                     & 10 & 0.9909 & 1.0 & 0.9894 & 1.0 & 0.7210 & 1.0\\
        \midrule
        ReL, $\alpha$=0.1 & 1 & 0.9971 & 0.9984 & 0.9986 & 0.9998 & 0.7774 & 0.9702\\
                     & 5 & 0.9930 & 0.9994 & 0.9932 & 1.0 & 0.7845 & 0.9638\\
                     & 10 & 0.9913 & 0.9993 & 0.9891 & 0.9993 & 0.7519 & 0.9793\\
        \midrule
        ReL, $\alpha$=0.3 & 1 & 0.9985 & 0.9989 & 0.9993 & 0.9987 & 0.8342 & 0.9361\\
                     & 5 & 0.9949 & 0.9992 & 0.9992 & 0.9893 & 0.8062 & 0.9250\\
                     & 10 & 0.9909 & 0.9985 & 0.9987 & 0.9881 & 0.7933 & 0.9078\\
        \midrule
        ReL, $\alpha$=0.5 & 1 & 0.9980 & 0.9989 & 1.0 & 0.9832 & 0.8518  & 0.9011\\
                     & 5 & 0.9945 & 0.9993 & 0.9997 & 0.9711 & 0.8728 & 0.9034\\
                     & 10 & 0.9909 & 0.9993 & 0.9995 & 0.9695 & 0.8877 & 0.8843\\
        \midrule
        ReL, $\alpha$=0.7 & 1 & 0.9980 & 0.9988 & 0.9999 & 0.9730 & 0.8902 & 0.8392\\
                     & 5 & 0.9962 & 0.9990 & 0.9998 & 0.9543 & 0.8726 & 0.7531\\
                     & 10 & 0.9912 & 0.9987 & 0.9998 & 0.9623 & 0.8699 & 0.7111\\
        \midrule
        ReL, $\alpha$=0.9 & 1 & 0.9979 & 0.9990 & 1.0 & 0.9566 & 0.9102 & 0.7734\\
                     & 5 & 0.9938 & 0.9987 & 1.0 & 0.9523 & 0.8830 & 0.6517\\
                     & 10 & 0.9892 & 0.9984 & 1.0 & 0.9419 & 0.8627 & 0.6333\\
        \bottomrule
    \end{tabular}
\end{table*}

\subsection{Relational Learner}
\label{sec_rel}
The AbsAE is trained so that the encoder $q_{\phi}(z|x)$ and the decoder $p_{\theta}(x|z)$ provide a mapping between the raw data samples and the corresponding generative factors. This can be exploited by the ReL in order to efficiently learn relations between those factors. The ReL model (Figure \ref{fig_arch_rel}) is composed of the \emph{relational learner} sub-network $r_\psi(z |z_1, .., z_N)$ that, in the case of a binary relation, is trained according to the following objective:
\begin{align}
\label{eq_rel_obj}
\max_\psi \quad \mathcal{L}_{REL}(z, i, j, \psi, f)
= p_{f(i, j)}\left(z\right)
\end{align}
where $z \sim r_\psi(z |z_i, z_j)$, $z_i$ and $z_j$ representing encoded data samples belonging, respectively, to the $i$-th and the $j$-th gaussian of $p(z)$. The function $f: \mathbb{N} \times \mathbb{N} \rightarrow \mathbb{N}$ is any function with domain and range in $[1, N]$ that characterizes the specific relation to be learned. Note the the number of arguments of $\mathcal{L}_{REL}$ actually depends on the arity of the desired relation. For example, when considering images of natural numbers, assuming that a relevant factor of variation is the number identity, the \emph{sum} relation can be learned by setting $f(i, j) = i + j$. Note that thus the prior distribution $p(z)$ is used to guide the learning process of the ReL, encouraging the model to encode the result into the desired component of the GM. This training can be done without the need of using additional data, as new samples can be drawn directly from the corresponding components of $p(z)$.
Having the ReL to operate in a structured latent space yields several advantages. First, in the latent space, a relation between generative factors is directly translated into a relation between components of $p(z)$. This means that the ReL can easily identify values of the generative factors of a data sample $x$ just by checking which component of $p(z)$ is the most active for encoding $x$. Additionally, since $p(z)$ is known, it is always possible to associate a probability threshold $\alpha$ to each input sample and each model prediction. This is useful in several ways. For input samples, it allows to identify potentially adversarial samples that are too far-away from the empirical data distribution (i.e. samples that ends up getting encoded into very low-probability regions of the latent space). For model's predictions, it provides additional useful information about the confidence level of the predictions.

\section{Experiments}
\label{sec_experiments}
We designed a set of experiments in order to inspect the following questions:
\begin{enumerate}
\item How well the AbsAE is capable of correctly clustering the values of generative factors in the latent space?
\item How well the ReL is capable of manipulating the latent representation in order to implement the desired relations?
\item How much the learned representations can be considered disentangled?
\end{enumerate}
In the rest of this section we accurately describe the experimental setting. First, we give a detailed account of the preprocessing procedures that is common to all the experiments, Then, we describe the crucial design choices made to implement the specific experiments\footnote{The source code of the project is available at \texttt{\url{https://github.com/Andrea-V/Weak-Disentanglement}}.}. In the Appendix, we also provide additional details and experimental results.

\subsection{Datasets and Preprocessing}
\label{sec_datasets}
We consider three datasets: the newly introduced Hand-Written Formulas (HWF) dataset \cite{li2020closed}, and the well-known dSprites \cite{2017_higgins_beta-vae} and Shapes3D \cite{2018_kim_factor_vae} datasets. The HWF dataset contains images of hand-written math formulas, consisting of the ten digits and three basic math operators. The dSprites dataset contains images of various 2-dimensional shapes, in different positions, scales and orientations. The Shapes3D dataset contains images of various 3-dimensional shapes in different colors combinations of (floor, shape, background) and rotations. In the case of HWF, the only relevant generative factor considered is the \emph{digit/operator identity}, for a total of 13 values (10 digits plus the sum, subtraction and multiplication operators), while everything else is considered a nuisance. For dSprites, we keep 3 values for the \emph{horizontal position} (left, center, right), 3 values for the \emph{vertical position} (up, center, down), and 3 values for the \emph{shape} (ellipse, square, heart). The \emph{scale} and \emph{orientation} are nuisances factors.
Finally, for Shapes3D, we keep 10 values for \emph{object color}, 4 values for \emph{shape} (cude, sphere, cylinder, ellipsoid) and 
3 values for \emph{scale} (small, medium, big), while considering \emph{floor color}, \emph{background color} and \emph{orientation} nuisance factors.
Thus, we end up having 13 factor combinations for HWF, 27 for dSprites and 120 for Shapes3D. Each of these combinations is represented as a single gaussian of the prior distribution $p(z)$.
Regarding relations, on the HWF dataset we consider the \emph{sum}, \emph{subtraction} and \emph{multiplication} binary relations. For dSprites, we consider 5 relations: \emph{move\_left}, \emph{move\_right}, \emph{move\_up}, \emph{move\_down} and \emph{change\_shape}. In Shapes3D, 5 relations are considered as well: \emph{+\_hue}, \emph{-\_hue}, \emph{change\_shape}, \emph{+\_scale}, \emph{-\_scale}.

In the case of HWF, the relations are chosen in order to reflect our intuitive understanding of the corresponding math operators. In dSprites and Shapes3D, on the other hand, the chosen relation have the effect of changing the value of a single factor of variation, while leaving the others unchanged. No restriction is imposed on the nuisance factors, that are able to vary freely when applying relations on the latent codes. We also perform data augmentation, corrupting the data samples by adding either bernoullian or gaussian noise to the original images. We split each dataset in training, validation and test set. The validation set is used to select the best values of the hyperparameters of the models, while the test set is used to compute the final results. The validation set is created by taking 10\% of the available training set. Similarly, the test set is created from 20\% of the total available data.

\begin{figure*}
     \centering
         \includegraphics[width=\textwidth]{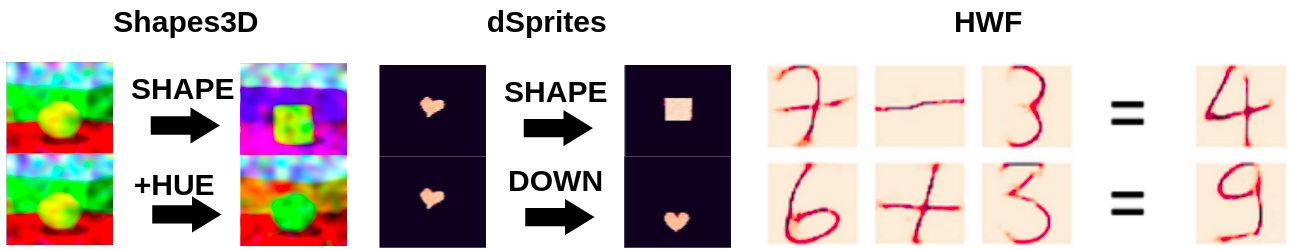}
        \caption{Samples of relations learned by the ReL on different datasets (additional samples are available in the Appendix).}
     \label{tab_relations}
\end{figure*}

\subsection{Training Settings}
\label{sec_training}
The encoder and decoder modules of AbsAE are implemented as a multilayer CNN architecture Additional details about the chosen architecture are contained in the Appendix. The hyperparameter have been tuned using trial and error, selecting the combination yielding the best performance on the validation set. All the networks used in the experiments are deterministic, i.e. $q_\phi(z|x)$ and $p_\theta(x|z)$ are Dirac's delta functions. The discriminator and the relational networks are implemented as a 3-layer MLP with 1024 units each. The hidden neurons use hyperbolic tangent non-linearities, while the output neurons use the sigmoid. In the experiments, we set the number of latent factors $N_z=8$ for HWF and dSprites, $N_z=16$ for Shapes3D. All tasks use a batch size of 1024 for the AbsAE's training and 128 for the ReL's training. We use the Adam optimizer with learning rate of $10^{-4}$ for HWF and dSprites, $10^{-5}$ for Shapes3D.

Initially, the training starts in a \emph{warmup phase}, were only the AbsAE is active. In this phase we set $p(z) \sim \mathrm{Uniform}(-1, 1)$, to encourage the latent codes to spread evenly across the latent space. During this phase only the AbsAE is trained. After 1000 epochs (5000 for Shapes3D), the \emph{full training phase} begins: the prior distribution is changed to the GM prior $p(z) \sim \frac{1}{N}\sum_{i=0}^N{\mathcal{N}(\mu_i, \Sigma_i)}$ described in Section \ref{sec_absae}. 
%We use 20 labelled samples for each generative factor combination to estimate the parameters $\mu_i$ and $\Sigma_i^2$ of the prior (30 samples for Shapes3D).
In this phase we also start the training of the ReL: the first step is to construct a training sample with the following structure:
\begin{equation}
    (z_{in_1}, ..., z_{in_R}, z_{rel}, z_{out})
\end{equation}
where $z_{in_1}, ..., z_{in_R}$ are the input latent codes of the relation, $R$ is the arity of the chosen relation, $z_{out}$ is the target latent code, and $z_{rel}$ is a code that identifies the relation. $z_{rel}$ can either be a symbolic code (such a categorical variable) or a latent code representing the specific relation.
Therefore, a training sample for the HWF dataset can be $(z_{2}, z_{3}, z_{+}, z_{5})$, where $z_{2}$ and $z_{3}$ are sampled from the prior's components corresponding to the digits ``2" and ``3", $z_{+}$ is sampled from the ``+" component, and $z_{5}$ is sampled from the ``5" component. On the other hand, when training the \emph{move\_up} relation on the dSprites dataset, a possible training sample will have the form $(z_{(\mathrm{center}, \mathrm{center}, \mathrm{square})}, z_{\mathrm{move\_up}}, z_{(\mathrm{center}, \mathrm{up}, \mathrm{square})})$, where $z_{(\mathrm{center}, \mathrm{center}, \mathrm{square})}$ is obtained by sampling from the prior component corresponding to the factor combination \{x\_position=\emph{center}, y\_position=\emph{center}, shape=\emph{square}\}, $z_{(\mathrm{center}, \mathrm{up}, \mathrm{square})})$ is sampled from the gaussian corresponding to \{x\_position=\emph{center}, y\_position=\emph{up}, shape=\emph{square}\}, and $z_{\mathrm{move\_up}}$ is a categorical variable that identifies the \emph{move\_up} relation. Unlike HWF, the dSprites dataset does not contain a way to identify the relations directly in the data, hence the need for an additional categorical variable for encoding relations.

Thus, the ReL learns how to perform changes to the latent codes from the starting region of the latent space to another one, according to the specific relation. Note that the training of the ReL can be done without the need of additional data, as the training samples can be constructed by directly sampling from $p(z)$. The elements of the training tuple are then concatenated together and sent in input to the ReL. We alternate a training iteration of the AbsAE with a training iteration of the ReL in order to learn both objectives at the same time. 
Training is carried on for 5000 more epochs (10000 for Shapes3D), for a total of 6000 epochs (15000).

%Figure \ref{fig_abs_samples} shows the decoded means of the learned gaussian priors for the different datasets, as well as some samples drawn from the individual gaussians.
%Finally, in Tab.\ref{tab_clustering_accuracy} we compare the clustering accuracy (i.e. the probability that a real-world instance is recognized as belonging to the right symbolic entity) of our model with the results reported in the recent work of \cite{ding2019clustering}.

\subsection{Structure of the Latent Space}
\label{sec_exp_absae}
The first set of experiments is meant to assess the capability of the prior distribution $p(z)$ to effectively identify and cluster the relevant regions of the latent space that identify a particular combination of generative factors. In Figure \ref{tab_clustering} are reported the clustering accuracy and the accepted ratio of the AbsAE on the test set of HWF, dSprites and shapes3D datasets. The results are obtained by first encoding a test sample $x$ to get its latent representation $z$. The classification is then performed by selecting the prior component that is more likely to have generated $z$. We report the results for different values of $\alpha$, that is, we compute the accuracy only on the test samples that reach a certain probability threshold $\alpha$ for at least one component of $p(z)$. If a sample does not reach the desired probability for any component, it is rejected, and the classification is not performed. We also report the ratio of test samples that the model does not reject (i.e. the \emph{acceptance ratio}) for each $\alpha$ threshold. A high accepted ratio means that a high proportion of the test samples has a high probability under the prior. 

The results in Figure \ref{tab_clustering} show that the clustering accuracy increases as $\alpha$ gets higher. For the HWF dataset, the model reaches over 90\% accuracy for each $\alpha$ thresholds. In the dSprites dataset, it takes longer to exceed 90\% accuracy, but the acceptance ratio stays very high for each $\alpha$ thresholds, meaning that the model is quite confident in its classifications. Shapes3D is, perhaps unsurprisingly, the most challenging dataset. Nevertheless, the AbsAE is still able to reach 90\% clustering accuracy for $\alpha \ge 0.7$, while keeping an acceptance ratio of over 75\%. We compare our results on the HWF dataset with the work of \cite{li2020closed}. Despite the more challenging setting (as the AbsAE is a generative model, our representations needs to also keep all the information needed for a good reconstruction the original data, whereas \cite{li2020closed} are only concerned with symbol classification) our model yields better results, obtaining higher accuracies for any values of $\alpha$.

\subsection{Manipulation of Latent Codes}
\label{sec_exp_rel}
The second set of experiments has the goal to test how well the ReL is capable of manipulating the learned latent representations in order to implement the desired relations. The relation accuracy of the model is computed by first sampling a latent code $z_{in}$ (or two, in the case of the binary relations of HWF) from the prior $p(z)$. After that, we choose a random relation $z_{rel}$ among the one that are available for that dataset and we feed both the $z_{in}$s and $z_{rel}$ in input to the ReL. If the \emph{depth} parameter is more than 1, we repeat this process accordingly, using the output of the ReL at the current step as input for the next step. The final output of the ReL $z_{out}$ is then classified by selecting the component of the GM prior with the highest probability of having generated $z_{out}$. Results are reported in Figure \ref{tab_relational}. We take into consideration different $\alpha$ thresholds and different depths of the relations.

The results shows that the performance of the ReL are only marginally affected by the specific $\alpha$ thresholds. There is a general tendency for the accuracy to increase, and the accepted ratio to decrease, as $\alpha$ get higher, but this mainly happens on the more challenging Shapes3D dataset. In the case of HWF and dSprites datasets, the accuracy stays at around 99\% and the accepted ratio is above 95\% for different values of $\alpha$ and different depths. This is a sign that the ReL can reliably learn the desired relations with high accuracy. The performance do not seem to be much affected by \emph{depth} parameter, meaning that the ReL is able to applying in cascade more than one relation to the same latent code without losing accuracy. Therefore, the learned relations can be combined compositionally in order to perform complex transformations of the initial latent representation. In Figure \ref{tab_relations} we report qualitative samples obtained from the ReL on different datasets.
%(additional samples are available in Appendix \ref{app_additional}). In Appendix \ref{app_more_tables} we report the data used to construct Figure \ref{fig_rel_results}.

\begin{table}
   \centering
   \caption{Disentanglement scores of latent representations on different datasets.}
   \label{tab_disentanglement}
    \begin{tabular}{c|ccc|ccc}
        \toprule
         & \multicolumn{3}{c}{\bf dSprites} & \multicolumn{3}{c}{\bf Shapes3D} \\
         & DCI & MIG & SAP & DCI & MIG & SAP \\
        \midrule
        $\beta$-VAE & 0.4566 & 0.602 & 0.67 & 0.153 & 0.270 & 0.131\\
        FactorVAE   & 0.8942 & 0.98  & 0.61 & 0.371 & 0.370 & 0.402\\
        \midrule
\cite{2019_locatello_few-labels} & 0.533 & 0.01 & 0.01 & 0.48 & 0.05 & 0.08 \\
 %LORD (CITE) & 0.4133 & 0.06 & 0.10 & 0.5933 & 0.18 & 0.43 \\
    \cite{gabbay2021image} & 0.8366 & 0.14 & 0.57 & \textbf{1.0} & 0.3 & \textbf{1.0} \\
        \midrule
        Ours & \textbf{0.9543} & \textbf{0.994} & \textbf{0.7728} & 0.6921 & \textbf{0.6897} & 0.5007\\
        \bottomrule
    \end{tabular}
\end{table}

\begin{figure*}
     \centering
         \includegraphics[width=\textwidth]{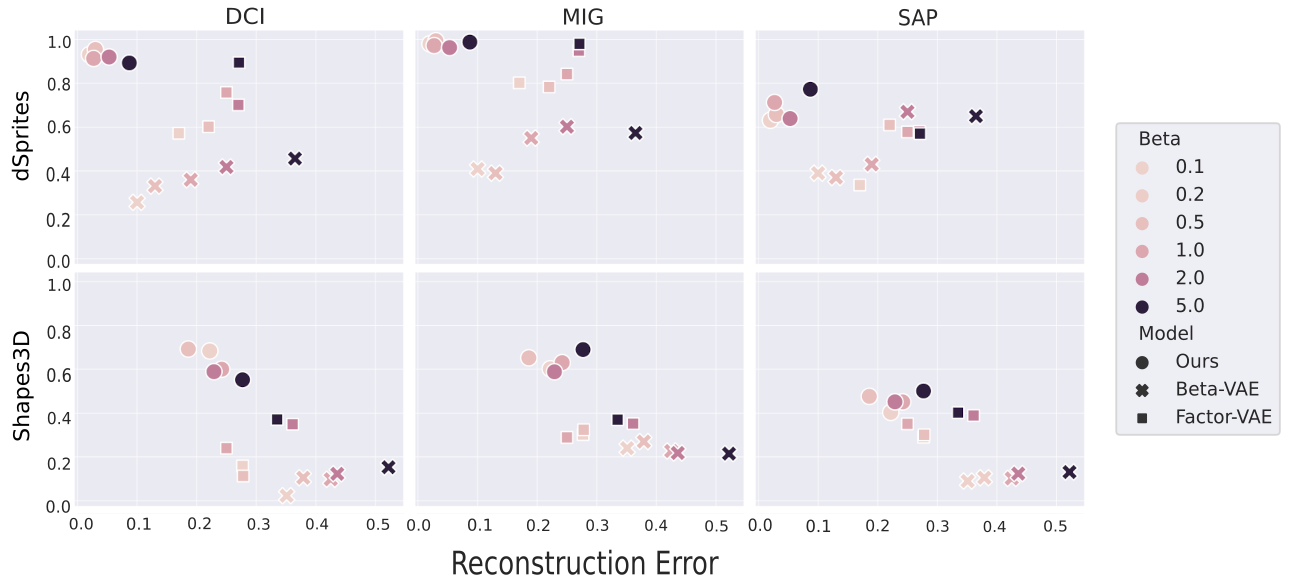}
        \caption{Disentanglement/reconstruction trade-off of the models on the considered datasets. The disentanglement metric (y-axis) is plotted against the reconstruction error (x-axis).}
     \label{fig_disentanglement}
\end{figure*}

\subsection{Disentanglement of Learned Representations}
\label{sec_exp_disentanglement}
Lastly, we wish to investigate how much the representations learned by our model can be considered disentangled by the ``classic" standards, while still keeping all the information needed to reconstruct the original data sample.
We trained a $\beta$-VAE \cite{2017_higgins_beta-vae} and FactorVAE \cite{2018_kim_factor_vae} on both the dSprites and Shapes3D datasets for comparison (the HWF dataset is not used, as disentanglement can only be measured when there are two or more relevant factors of variation in the data). In order to ensure comparability of results, we use the same encoder-decoder architectures. We also keep the same learning rates and train for the same number of epochs as the one described in \ref{sec_training}. We then repeats the experiments for different values of $\beta$ (note that, in FactorVAE, we consider $\beta$ to be the hyperparameter controlling only the total correlation term of the loss function).
The ideal model would score highly in the various metric, while keeping the reconstruction error as low as possible for different values of $\beta$. Hence, the better models are the ones ending up in the upper-left region of the plots. In Figure \ref{fig_disentanglement} we plot the scores of popular disentangled metrics: SAP score \cite{2017_kumar_variational}, MIG score \cite{2018_chen_beta-TCVAE}, and DCI score \cite{2018_eastwook_framework} against the reconstruction accuracy of our model on the dSprites and Shapes3D datasets (in the case of DCI, we report the average of \emph{disentanglement}, \emph{completeness} and \emph{informativeness} scores). Since the above metrics assume to receive an input representations that follows the classic notion of disentanglement (i.e. where each individual dimension is responsible for encoding a single factor of variation), we transform each latent codes into its corresponding generative factors before computing the metrics for our model. This step can be done efficiently, as all the information about the factor of variations can be inferred just by classifying the latent code as described in Section 
\ref{sec_rel}.

The results  of Figure \ref{fig_disentanglement} shows that our model's representations offer the best tradeoff between reconstruction and disentanglement. Our representation overall is not losing much reconstruction information as $\beta$ is increased. On the other hand, the $\beta$-VAE's disentanglement scores can only be improved at the expense of reconstruction error, which begin to increase quickly as $\beta$ becomes larger. FactorVAE's representation ends up in between, still not able to reach the same disentanglement/reconstruction tradeoff as our model. Finally, in Table \ref{tab_disentanglement} we directly compare the disentanglement performance of our model against the state of the art works of \cite{2019_locatello_few-labels} and \cite{gabbay2021image}, as well as the $\beta$-VAE and FactorVAE baseline models. The results show that our model is able to reach superior disentanglement performance on the dSprites dataset, while still being competitive with the other state-of-the-art models on Shapes3D. Overall, it seems that our disentanglement disentanglement approach is able to identify and encode the relevant factors of variation without affecting the reconstruction power of learned representations.

\section{Conclusion}
\label{sec_conclusion} 
We proposed the \emph{weak disentanglement} meta-prior, a method for implementing disentanglement of latent representations of generative models by leveraging additional relational information. We presented a new generative model that implements our approach, divided into an auto-encoding part (AbsAE) and a relational learning part (ReL).
We tested our approach on three different datasets of increasing complexity. The experiments shows that the AbsAE is able to identify and isolate the relevant regions of the latent space with high accuracy. The ReL is able to correctly manipulate the latent representations, even when applying multiple relations in sequence on the same representation. Finally, the learned representations yields better disentanglement scores when tested against similar models that rely on the ``classic" notion of disentanglement, while preserving the information needed to achieve a good reconstruction of the original data sample, showing that our approach can be a viable option for disentanglement. The imposed structure of the latent space makes the model robust to potentially adversarial sample, as well as providing additional information about the confidence of individual predictions.
%We believe that weakly disentangled representations can be a practical approach to the disentanglement problem and help to  push forward the research in the field.
In the future, we plan to further refine the structure of latent space learned by the AbsAE. It could be useful to encode the generative factors of the data in different latent spaces to encourage modularity of learned representations and to prevent the number of gaussian component of $p(z)$ to become too large when the number of values of generative factors increases. Another research direction is in finding a more expressive prior distribution $p(z)$. We also plan to enhance the ReL's overall architecture, possibly employing a graph neural network \cite{bacciu2020gentle} to learn more expressive relations over the data.

%\section*{Acknowledgment}

\bibliographystyle{IEEEtran}
\bibliography{IEEEabrv, bibliography}

\appendix
We strongly encourage the interested readers to check the additional material, available at the following URL:\\ \texttt{\url{https://ufile.io/1vpsoujp}}.

The additional material contains:
\begin{itemize}
    \item A detailed descriptions of the AbsAE and ReL architectures.
    \item Qualitative samples of the AbsAE's latent space.
    \item Additional qualitative samples of relations learned by the ReL.
\end{itemize}

\end{document}